\documentclass[runningheads]{llncs}
\usepackage{cite}
\usepackage{amsmath,amssymb,amsfonts}
\usepackage{algorithmic}
\usepackage[ruled]{algorithm2e}
\usepackage{graphicx}
\usepackage{textcomp}
\usepackage{multirow, multicol}
\usepackage{placeins}
\usepackage{xcolor, soul}
\usepackage{graphicx}
\usepackage{multirow}
\usepackage{booktabs}
\usepackage{tabularx}

\usepackage{subcaption}
\begin{document}
\setlength{\textfloatsep}{5pt}
\title{Enhancing Sentiment Analysis through Multimodal Fusion: A BERT-DINOv2 Approach}
 \author{Taoxu Zhao\inst{1} \and
 Meisi Li\inst{1} \and
 Kehao Chen\inst{1} \and
 Liye Wang\inst{1} \and
 Xucheng Zhou \inst{1} \and
Kunal Chaturvedi\inst{2} \and
 Mukesh Prasad\inst{2} \and
 Ali Anaissi\inst{1,2} \and
 Ali Braytee\inst{2}}
 \authorrunning{T. Zhao et al.}

 \institute{The University of Sydney, Camperdown, Australia \and
 University of Technology Sydney, Ultimo, Australia} 
% \email{ali.braytee@uts.edu.au}\\

\maketitle              % typeset the header of the contribution
\begin{abstract}
Multimodal sentiment analysis enhances conventional sentiment analysis, which traditionally relies solely on text, by incorporating information from different modalities such as images, text, and audio. This paper proposes a novel multimodal sentiment analysis architecture that integrates text and image data to provide a more comprehensive understanding of sentiments. For text feature extraction, we utilize BERT, a natural language processing model. For image feature extraction, we employ DINOv2, a vision-transformer-based model. The textual and visual latent features are integrated using proposed fusion techniques, namely the Basic Fusion Model, Self-Attention Fusion Model, and Dual-Attention Fusion Model. Experiments on three datasets—the Memotion 7k dataset, MVSA-single dataset, and MVSA-multi dataset—demonstrate the viability and practicality of the proposed multimodal architecture.
%Multimodal sentiment analysis enhances conventional sentiment analysis, which traditionally relies only on text, by incorporating information from multiple modalities such as image, text and audio. This paper proposes a multimodal sentiment analysis architecture integrating text and image data to understand sentiments comprehensively. For text feature extraction, we utilize BERT, a state-of-the-art model in natural language processing. For image feature extraction, we use DINOv2, a vision-transformer-based model. The textual and visual latent features are integrated using the proposed fusion techniques, i.e., Basic Fusion Model, Self-Attention Fusion Model, and Dual-Attention Fusion Model. The experiments on the three datasets (Memotion 7k dataset, MVSA-single dataset and MVSA-multi dataset) demonstrate the viability and practicality of the proposed multimodal architecture. 
\end{abstract}

\section{Introduction}
\label{sec:introduction}

\begin{figure*}
    \centering
    \includegraphics[width=0.8\textwidth]{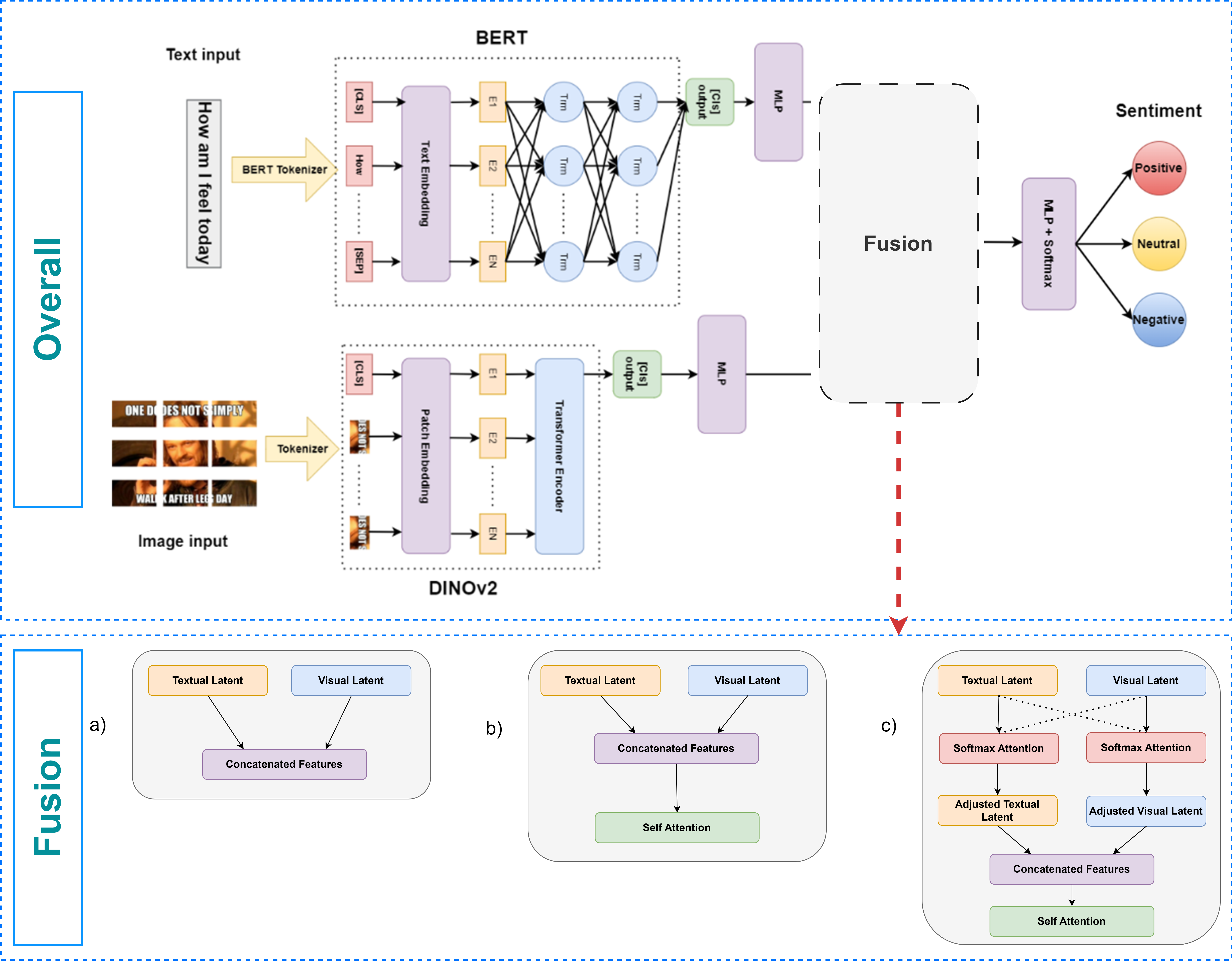}
    \caption{The overall architecture of the proposed framework (Above). The fusion methodology of the framework (Below): a) Basic Fusion Model; b) Self-Attention Fusion Model; c) Dual-Attention Fusion Model}
    \label{overallarch}
\end{figure*}

% Sentiment analysis is a sub-field of natural language processing (NLP) \cite{RODRIGUEZIBANEZ2023119862}. Traditionally, sentiment analysis relies on statistical models to parse and interpret sentiments from specified inputs. However, the recent surge in the diversity of data types on social media platforms – including image, video, and audio, sometimes accompanied by textual captions – has transformed the nature of inputs in sentiment analysis. This evolution marks a shift from simplistic unimodal data (solely text-based) to increasingly complex multimodal data. Consequently, this transition necessitates adapting statistical models to multimodal scenarios to cater to these emerging requirements \cite{GANDHI2023424}. 

Multimodal sentiment analysis works process data and combine the different data modalities first, then use a single model to analyse the combination of feature vectors of the data input, but lack the consideration of the correlation between data modalities. For example, facial and other body features can be implemented with the speaker’s words, and different facial expressions with the same words can have different meanings. Recently, many multimodal sentiment analysis approaches have been proposed. The authors \cite{sentistrength} integrate SentiBank's capability to extract 1200 adjective-noun pairs for image classification with SentiStrength's text-based sentiment scoring \cite{sentibank}. CNN-Multi \cite{kim2014convolutional} leverages pre-trained CNNs for text and image feature extraction, concatenating these features for subsequent classification through four fully connected layers. Similarly, DNN-LR \cite{DNN_LR} utilizes pre-trained convolutional neural networks (CNNs) for extracting text and image representations, employing logistic regression for sentiment classification. MultiSentiNet's approach \cite{MVSA_paper}, combines long short term memory (LSTM) model and visual geometry group (VGG) model for feature extraction, integrating results via weighted average or image feature-guided attention. 

The enhancement of unimodal statistic models, which can serve as components within multimodal frameworks, is progressing through the adoption of innovative methodologies. Notably, the emergence of Bidirectional Encoder Representations from Transformers (BERT) \cite{devlin2019bert} and its variants have demonstrated superior performance over preceding models, such as recurrent neural networks (RNNs) and LSTMs in text-based sentiment classification due to its refined understanding of textual syntax and semantics. In parallel, the introduction of Vision Transformers \cite{dosovitskiy2021image}, represents a significant extension of the transformer model architecture, originally conceived for NLP applications, into computer vision. These models leverage the strengths of large pre-trained datasets, showing enhanced efficacy than CNNs in many image-related tasks. However, we find that vision transformer models may be able to better deal with the vision feature than CNNs.

A good latent representation greatly influences the sentiment analysis results and provides an excellent foundation for the following fusion phase. Therefore, our research aims to harness the capabilities of these newly advanced techniques through their integration via innovative fusion methodologies.  Modal fusion is one of the main challenges in multimodal constructions. A single concatenation method \cite{vo2019composing} can effectively integrate the multimodal information. Huang et al. \cite{Huang2019imagetext} proposed a model using a "Deep Multimodal Attentive Fusion" method to deal with the fusion issue. They also pointed out the drawbacks of the early fusion methods and emphasized the great performance of late fusion methods as they cannot always unlock the full potential of each modal data. Yu et al. \cite{Yu2019AdaptingBF} used the attention technique adding on the BERT model to deal with target-oriented tasks. Tsai et al. \cite{tsai2019multimodal} proposed cross-attention technique to combine the different latent representations effectively, enhancing the performance of each single modality. Recently, Lee et al. \cite{lee2024modeling} explored multimodal learning by leveraging BERT and DINOv2 for modeling social interactions. However, their focus is on aligning language-visual cues for referent tracking and speaker identification, whereas our work aims at sentiment analysis and proposes novel attention-based fusion techniques tailored for this specific task. The methodologies presented in these papers have guided us in establishing a late fusion strategy and have inspired us to employ attention-based fusion methods.

We select two advanced unimodal models: BERT \cite{devlin2019bert} and DINOv2 \cite{oquab2023dinov2} (a vision transformer-based model) and then propose three fusion methods to create a multimodal (textual and visual) sentiment analysis model combining the extracted features from separate (text and image) modalities, named Basic Fusion Model, Self-Attention Fusion Model, and Dual-Attention Fusion Model. These methods are designed to synergize the distinct features extracted from separate modalities. 

Our contributions:
\begin{itemize}
    \item Combined extracted features from multimodal data using different fusion methods, such as the Basic Fusion Model, Self-Attention Fusion Model, and Dual-Attention Fusion Model.
    \item Extensive experiments to compare the effectiveness of the methods using three multimodal sentiment analysis datasets.

\end{itemize}
 
\section{Method}
Here, first, we explain the unimodal information extraction process for both textual and visual context information, respectively. Next, we detail the various fusion methods that combine latent features from both text and image data to be processed through several fully connected layers to predict sentiment analysis. The overall framework is shown in Figure \ref{overallarch}.
\subsection{Textual Features}
BERT \cite{devlin2019bert} is an existing innovative model meticulously designed for a series of natural language understanding tasks. BERT's strength is its pre-training tasks and bidirectional representation, which make it well-known for various downstream NLP tasks. Traditional models like LSTM read the text input in one direction, either forward or backward, which makes them less effective in understanding the contextual relationship between words \cite{Manasa2024}. BERT's bidirectional attention mechanism allows it to consider both the preceding and following text, offering a more comprehensive understanding of each word within its context \cite{devlin2019bert}.

This pre-training involves tasks such as Next Sentence Prediction (NSP) and Masked Language Modeling (MLM), which equip the model with a strong understanding of language semantics and structure \cite{devlin2019bert}. As a result, the pre-trained BERT model can serve as a robust starting point for a wide range of NLP tasks, from text summarization to question answering. Given BERT's advantages, it's evident that its capability to understand textual context is excellent. Therefore, it is well-suited to serve as our text feature extraction model. In the proposed architecture, we utilize a BERT layer for initial text processing. This involves feeding a sequence of input tokens \( X_t = \{x_1, x_2, ..., x_n\} \) into the BERT model, which produces a sequence of output embeddings \( H = \{h_1, h_2, ..., h_n\} \). Each embedding \( h_i \) is a 768-dimensional vector that captures the contextual information of the corresponding token within the entire sequence.

To adapt these embeddings for our multimodal sentiment analysis, we apply a linear transformation directly to the entire sequence of output embeddings from BERT, reducing the dimensionality of each vector from 768 to 256. The transformation defined in Eq. \ref{trans} is designed to condense the rich linguistic information encoded by BERT into a more compact and relevant representation for further analysis.

\begin{equation}
\label{trans}
    t_i = Wh_i + b
\end{equation}

where \( W \) is a \( 768 \times 256 \) weight matrix, \( b \) is a bias vector, and \( t_i \) represents the transformed embedding of the ith token.

For the final textual latent representation \( t \), defined in Eq. \ref{deft}, we use the transformed CLS embedding \( t_{\text{CLS}} \), which represents a condensed view of the entire input sequence:

\begin{equation}
\label{deft}
    t = t_{\text{CLS}}
\end{equation}
This approach leverages the CLS token's embedding directly after merging all the necessary information for sentiment analysis following its transformation. The representation, \( t \), is then used as part of the fusion process.

\subsection{Visual Features}
DINOv2 \cite{oquab2023dinov2} is the name of a series of pre-trained visual models, which are trained using Vision Transformers (ViT) \cite{dosovitskiy2021image} on  LVD-142M dataset. The key distinction between the ViT and traditional transformer architectures is that ViT divides an image into individual patches and treats these patches as tokens. In this way, the image is viewed as a sequence of tokens. ViT processes the entire image as a unified sequence, allowing it to capture global information instead of CNNs, focusing only on local information through the convolutional kernels. As a result, ViT often outperforms CNNs in various scenarios.

However, when referring to DINOv2, it not only refers to the ViT models but also a training methodology, a self-supervised learning approach \cite{oquab2023dinov2}. Self-supervised learning achieves excellent segmentation results and robust performance, without fine-tuning, across various computer vision tasks. DINOv2 can learn from any collection of images and acquire features that current standard methodologies can't learn, making it one of the most advanced visual models today.

In the architecture, DINOv2 serves as the foundational layer for extracting complex features directly from image inputs. Each input image \( I \) is first divided into a sequence of patches \( \{p_1, p_2, ..., p_m\} \), which are then linearly embedded into tokens and processed through the DINOv2 transformer network. This process (see Eq. \ref{dinov2pr}) transforms each image into a sequence of patch embeddings, providing a rich representation of the visual information:

\begin{equation}
\label{dinov2pr}
    G = \text{DINOv2}(I) = \text{DINOv2}(p_1, p_2, ..., p_m)
\end{equation}

Upon processing images through the transformer, the proposed model employs a custom linear transformation layer, which maps the transformer's output from a 384-dimensional space to a 256-dimensional space for each patch embedding. This transformation is designed to adapt the broad capabilities of the DINOv2 by refining the feature representations to a more manageable size for the analysis of sentiments:

\begin{equation}
    v_i = Wg_i + b
\end{equation}
where \( W \) is a \( 384 \times 256 \) weight matrix, \( b \) is a bias vector, and \( v_i \) represents the transformed embedding of the ith patch.

For the final visual latent representation \( v \), defined in Eq. \ref{finCon}, we use the transformed global feature embedding \( v_{\text{CLS}} \), which represents a condensed view of the entire image:
\begin{equation}
\label{finCon}
    v = v_{\text{CLS}}
\end{equation}
This approach leverages the global feature embedding directly. It is then used as part of the fusion process with textual features extracted from the BERT model in our sentiment analysis framework.

% \begin{figure}
%     \centering
%     \includegraphics[width=0.45\textwidth]{models/Fusion.png}
%     \caption{The fusion methodology of the framework. Here a)Basic Fusion Model; b)Self-Attention Fusion Model; c)Dual-Attention Fusion Model}
%     \label{fusionarch}
% \end{figure}

\subsection{Attentional Feature Fusion}

The proposed architecture integrates and leverages the strengths of both textual and visual representation models for multimodal sentiment analysis. The integration allows for the extraction and fusion of high-level features from both text and images, facilitating a comprehensive understanding of multimodal data. For this, we propose three fusion methods. A Basic Fusion Model that concatenates the textual and visual latent representation (as shown in Figure \ref{overallarch} a). Let \( t \) be the textual latent representation obtained from the BERT model, and \( v \) be the visual latent representation obtained from the DINOv2 model. Both \( t \) and \( v \) are vectors. The simple concatenation \( c \) is defined in Eq. \ref{concat_formula}

\begin{equation}
\label{concat_formula}
    c = [t, v]
\end{equation}
where \( [t, v] \) represents the concatenation of vectors \( t \) and \( v \). The dimension of \( c \) will be the sum of the dimensions of \( t \) and \( v \). To capture the mutual information of the concatenated latent representation, we introduce a self-attention layer after the concatenation layer as the second method, namely the Self-Attention Fusion Model (as shown in Figure \ref{overallarch} b). The output \( s \) of the self-attention mechanism is defined in Eq. \ref{selfatt} as,

%\FloatBarrier

\begin{equation}
\label{selfatt}
    s = \text{Softmax}\left(\frac{cW_Q(cW_K)^T}{\sqrt{d_k}}\right) cW_V
\end{equation}
where $c$ is the concatenated vector given from Eq. \ref{concat_formula}. \( W_Q \), \( W_K \), and \( W_V \) are the weight matrices for the query, key, and value, respectively, which are applied to the vector c. \( d_k \) is the dimension of the key.

%\FloatBarrier

The third method is the Dual-Attention Fusion Model (as shown in Figure \ref{overallarch} c). Based on the second fusion method, it additionally uses information from another modality to adjust the latent representation vector of each modality, employing a cross-modal attention mechanism. First, queries, keys, and values for both modalities are computed using Eq. \ref{qkv},

\begin{equation}
    \label{qkv}
    \begin{aligned}
        Q_t &= tW_{Q_t}, & K_t &= tW_{K_t}, & V_t &= tW_{V_t}, \\
        Q_v &= vW_{Q_v}, & K_v &= vW_{K_v}, & V_v &= vW_{V_v}.
    \end{aligned}
\end{equation}

Next, we apply softmax attention for cross-modal adjustments as shown in Eq. \ref{softmaxatt},

\begin{equation}
    \label{softmaxatt}
    \begin{aligned}
        t' &= \text{Softmax}\left(\frac{Q_t K_v^T}{\sqrt{d_k}}\right) V_v \\
        v' &= \text{Softmax}\left(\frac{Q_v K_t^T}{\sqrt{d_k}}\right) V_t
    \end{aligned}
\end{equation}

Finally, we apply concatenation and self-attention as defined in Eq. \ref{finalconc},

\begin{equation}
\label{finalconc}
    \begin{aligned}
        s' &= \text{Softmax}\left(\frac{c'W_Q (c'W_K)^T}{\sqrt{d_k}}\right) c'W_V
    \end{aligned}
\end{equation}

where $c'$ is the concatenation of $t'$ and $v'$.

%\FloatBarrier

\section{Experiments}
\subsection{Datasets}

Memotion 7k Dataset \cite{memotion} consists of 6992 samples, each paired with an image and the corresponding caption, representing a complete 'meme'. The images are scraped using the Google Images search engine, and the annotation process is carried out with the help of five annotators. The final annotations are adjudicated based on a majority voting scheme. The dataset is part of a challenge that includes three subtasks: analyzing memes for sentiment, which can be positive, negative, or neutral.

MVSA Datasets \cite{MVSA} collected from X, are designed for sentiment analysis on multi-view social data. The MVSA-Single dataset contains 4,869 pairs of images and texts labelled by a single annotator. The MVSA-Multi dataset consists of 19,600 text-image pairs, each labeled by three annotators, ensuring a richer and more robust sentiment analysis. Similarly to Memotion 7k Dataset, MVSA datasets have three classes: positive, neutral, and negative.

\subsection{Experiment settings}

Specifically, for the Memotion dataset, we employed macro F1 as our evaluation metric, requiring us to pay attention to the performance across each class. For MVSA-single and MVSA-multi, we used accuracy and F1-score. We utilized focal loss \cite{lin2018focal} as the loss function and adjusted the \(\gamma\) parameter ($\gamma$=2, 3, 4) to ensure the model adequately focuses on the minority classes. However, for the MVSA dataset, our primary evaluation metrics are weighted F1 and accuracy. 

To improve the performance of our model, we tuned a set of hyperparameters to facilitate model convergence. Across all datasets, we experimented with different learning rates: 0.01, 0.001, 0.0001, and 0.00001. Adam optimizer is used with dropout rates set to 0, 0.2, and 0.5. Cross-entropy loss function has been used in the experiments.

% \subsection{Time complexity}

\section{Results and Discussion}

In this section, we present experiments conducted on the proposed fusion methods across three different datasets and compare them with benchmark models and state-of-the-art models so that the usability and generalizability of models can be ensured. Additionally, the ablation study is performed to provide in-depth research on the model's fusion methods. At last, we show experiments conducted while tuning hyperparameters for all models across all datasets.

\subsection{Comparison with the state of the art}
To better demonstrate the usability and SOTA performance of our proposed model in this domain, we compare it with the literature corresponding to the relevant datasets using the same experimental setup. 

\subsubsection{MVSA datasets}

The comparative analysis presented in Table \ref{MVSA_table} showcases the performance of various models on the MVSA-single and MVSA-multi datasets. Notably, the integration of BERT and DINOv2 models, through concatenation, achieves the best performance on the MVSA-single dataset, with an Accuracy of 0.73 and an F1 score of 0.71, surpassing all other models including MultiSentiNet's attention-based approach (MultiSentiNet-Att) \cite{MVSA_paper}, which leads on the MVSA-multi dataset with an accuracy of 0.68 and an F1 score of 0.68. 

This approach outperforms the previously established benchmarks, including SentiBank \cite{sentistrength}, CNN-Multi \cite{cai0_14}, DNN-LR models \cite{DNN_LR} and HSAN \cite{8004895}, and even the advanced MultiSentiNet  \cite{MVSA_paper} and Dual-Pipeline \cite{10.1145/3589335.3651967} models. The BERT+DINOv2 model with additional self-attention mechanisms also shows strong performance, underlining the potential of attention mechanisms in multimodal sentiment analysis. This comparison underscores the advances in multimodal sentiment analysis, demonstrating that the fusion of high-performing models like BERT and DINOv2, especially when combined with sophisticated techniques such as self-attention, can lead to substantial improvements. It is worth noting that our model falls short of the state of the art model in the MVSA-multi dataset, which warrants further discussion in the subsequent section. Overall, the performance of our model architectures remains strong, robust, and adaptable.

%Kunal: Missing Acc from Dual-Attention Fusion Model

\begin{table*}[t]
\caption{Results for MVSA-single and MVSA-multi Compared to the Literature}
\label{MVSA_table}
\centering
\begin{tabular}{|l|c|c|c|c|}
\hline
\textbf{Model} & \multicolumn{2}{c|}{\textbf{MVSA-single}} & \multicolumn{2}{c|}{\textbf{MVSA-multi}} \\
\cline{2-5}
 & \textbf{Acc} & \textbf{F1} & \textbf{Acc} & \textbf{F1} \\
\hline
SentiBank (image only) \cite{sentistrength}          & 0.45 & 0.43 & 0.55 & 0.51 \\
SentiStrength (text only) \cite{sentistrength}       & 0.49 & 0.48 & 0.50 & 0.55 \\
SentiBank+SentiStrength \cite{sentistrength}         & 0.52 & 0.50 & 0.65 & 0.55 \\
HSAN \cite{8004895}                                  & -    & 0.66 & -    & 0.67 \\
DNN-LR \cite{DNN_LR}                                 & 0.61 & 0.61 & 0.67 & 0.66 \\
CNN-Multi \cite{cai0_14}                             & 0.61 & 0.58 & 0.66 & 0.64 \\
MultiSentiNet-Avg \cite{MVSA_paper}                  & 0.66 & 0.66 & 0.67 & 0.66 \\
MultiSentiNet-Att \cite{MVSA_paper}                  & 0.69 & 0.69 & 0.68 & 0.68 \\
Dual-Pipeline \cite{10.1145/3589335.3651967}         & 0.57 & 0.56 & \textbf{0.73} & 0.69 \\
\textbf{Ours (Basic Fusion Model)}                   & \textbf{0.73} & \textbf{0.71} & 0.68 & 0.67 \\
\textbf{Ours (Self-Attention Fusion Model)}          & 0.72 & 0.70 & 0.68 & 0.67 \\
\textbf{Ours (Dual-Attention Fusion Model)}          & 0.72 & \textbf{0.71} & 0.67 & 0.66 \\
\hline
\end{tabular}
\end{table*}

%
% \begin{table}[H]
% \caption{Result for MVSA-single \& MVSA-multi compared to the literature}
% \label{MVSA_table}
% \centering
% \begin{tabular}{|c|cc|cc|}
% \hline
% \multirow{2}{*}{\textbf{Model}} & \multicolumn{2}{c|}{\textbf{MVSA-single}} & \multicolumn{2}{c|}{\textbf{MVSA-multi}} \\
% \cline{2-5}
% & \textbf{Acc} & \textbf{F1} & \textbf{Acc} & \textbf{F1} \\
% \hline

% SentiBank (image only) & 0.4522 & 0.4380 & 0.5502 & 0.5115 \\
% SentiStrength (text only) & 0.4986 & 0.4845 & 0.5057 & 0.5536 \\
% SentiBank+SentiStrength & 0.5205 & 0.5008 & 0.6562 & 0.5536 \\
% CNN-Multi & 0.6120 & 0.5837 & 0.6639 & 0.6419 \\
% DNN-LR & 0.6142 & 0.6103 & 0.6786 & 0.6633 \\
% HSAN & - & 0.6690 & - & 0.6776 \\
% MultiSentiNet-Avg & 0.6674 & 0.6659 & 0.6786 & 0.6649 \\
% MultiSentiNet-Att & 0.6984 & 0.6963 & \textbf{0.6886} & \textbf{0.6811} \\
% \hline
% Ours (Basic Fusion Model) & \textbf{0.7323} & \textbf{0.7198} & 0.6828 & 0.6735 \\
% Ours (Self-Attention Fusion Model) & 0.7201 & 0.7077 & 0.6871 & 0.6741 \\
% Ours (Dual-Attention Fusion Model) & - & 0.7190  & - & 0.6676 \\
% \hline
% \end{tabular}
% \end{table}

%

%%%%%%%%%%%%%%%%%%%%%%%%%%%%%%%%%%%
\subsubsection{Memotion 7k dataset}

As detailed in Table \ref{Memotion_table}, our approach is benchmarked against the competition baseline \cite{memotion} and the leading models from the leaderboard. The leading models from the competition, as highlighted in the literature, showcase a wide array of computational strategies. Vkeswani
IITK from Keswani et al. \cite{keswani2020iitk} exemplifies this diversity by employing a spectrum of algorithms, ranging from fundamental linear classifiers like FFNN and Naive Bayes to more sophisticated transformers such as MMBT \cite{rahman2020integrating} and BERT \cite{devlin2019bert}, applied across both text-only and combined text-image inputs. Similarly, Guo et al. \cite{guo2020guoym} utilizes an ensemble methodology, leveraging textual features extracted via Bi-GRU, BERT, or ELMo in conjunction with visual features sourced from Resnet50. The approach "Aihaihara" as mentioned in \cite{memotion} opts for a concatenated approach, merging visual cues from a VGG-16 pre-trained model with textual insights from an n-gram language model, illustrating the varied approaches towards integrating visual and linguistic data. Das and Mandal \cite{das2020team} employed a multi-task learning framework that combined ResNet18 for the visual task and bidirectional
LSTM and GRU for textual tasks. Bejan \cite{bejan2020memosys} combined the BERT model for text understanding and VGG16 pre-trained model for extracting visual information.

\begin{table}[h]
\centering
\caption{Result for Memotion 7k}
\label{Memotion_table}
\resizebox{0.45\textwidth}{!}{
\begin{tabular}{|c|c|}
\hline
\textbf{Model} & \textbf{Macro F1} \\
\hline
Competition Baseline \cite{memotion} & 0.2176 \\
Vkeswani IITK \cite{keswani2020iitk} & 0.3546 \\
Guoym \cite{guo2020guoym} & 0.3519 \\
Aihaihara \cite{memotion} & 0.3501 \\
Sourya Diptadas \cite{das2020team} & 0.3488 \\
MemoSYS \cite{bejan2020memosys}  & 0.3475 \\
Ours (Basic Fusion Model) & 0.3237 \\
Ours (Self-Attention Fusion Model) & 0.3436 \\
Ours (Dual-Attention Fusion Model) & \textbf{0.3552} \\
\hline
\end{tabular}
}
\end{table}

The results demonstrate that our model, leveraging a sophisticated fusion of softmax attention and self-attention mechanisms, not only surpasses the competition baseline but also edges out the top-performing models from the leaderboard with a macro F1 score of 0.3552. This achievement underscores the effectiveness of our model in handling the intricacies of multimodal sentiment analysis, showcasing its potential to advance the state of the art in this field.

\subsection{Breakdown Analysis and Discussion}

\begin{figure}[ht]
    \centering
    % First figure
    \begin{minipage}{0.45\textwidth}
        \centering
        \includegraphics[width=\textwidth]{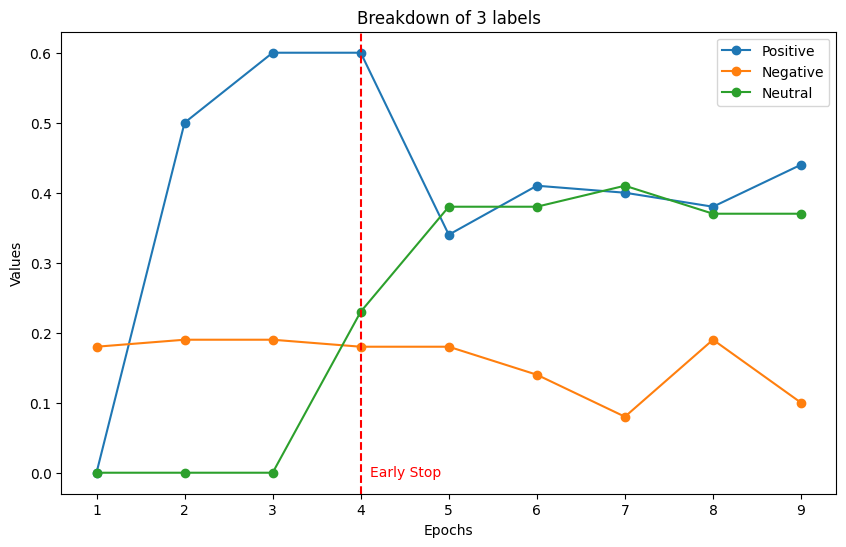}
        \caption{Breakdown performance for the best model for Memotion 7k dataset}
        \label{breakdown_memotion}
    \end{minipage}
    \hfill
    % Second figure
    \begin{minipage}{0.45\textwidth}
        \centering
        \includegraphics[width=\textwidth]{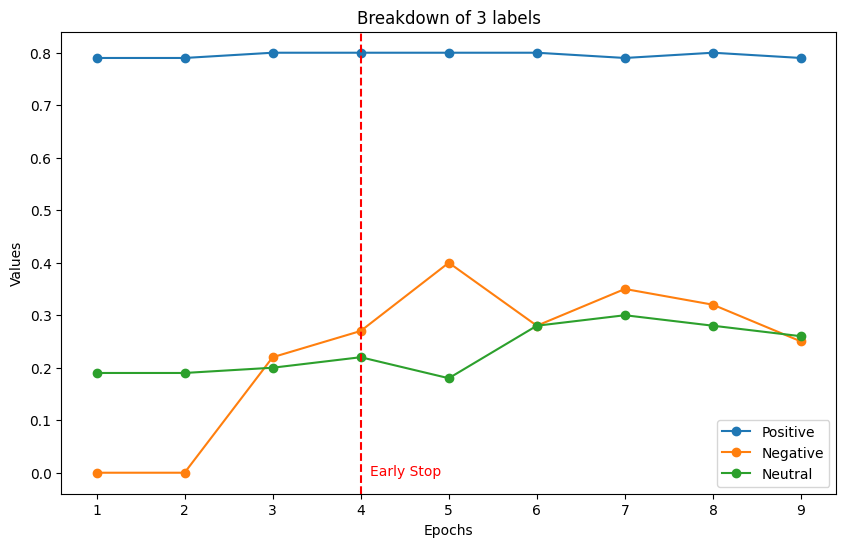}
        \caption{Breakdown performance for the best model for MVSA-multi dataset}
        \label{breakdown_mvsamulti}
    \end{minipage}
\end{figure}

Examining the breakdown performance plot (as shown in Figure \ref{breakdown_mvsamulti}) in the training process, we can easily discover the issue. Due to Adam's adaptive learning rate \cite{kingma2017adam}, the model converges well right from the start, especially for the positive class, reaching near-optimal performance almost in the first epoch. The bias towards predicting the positive class helps maintain high levels of weighted F1 and accuracy. However, for the neutral and negative classes, the model learns fewer relevant features by comparison. From the cases in MVSA-multi, we have come to realize that in addition to understanding the overall performance of the model, the predictive capability for each class is also crucial. 

However, the evaluation criteria for the Memotion dataset and the MVSA dataset differ; Memotion focuses on predictions for each class, while MVSA emphasizes overall performance. Therefore, in this section, we aim to explore the predictive performance for different classes, and using the Memotion dataset is a good choice for this purpose. For further investigating how the best model maintains its learning for each class during the model convergence process, we monitored the changes in the F1 score for each class (as shown in Figure \ref{breakdown_memotion}).

%\FloatBarrier
From Figure \ref{breakdown_memotion}, in the epoch where early stopping occurs, we find that the model still inevitably achieves the highest F1 score on the positive class, as it is the dominant class. However, we can see that although there is a large difference in the number of samples between the neutral and negative classes, their predictive performance is similar and not far off from that of the positive class. This suggests that focal loss has effectively captured the minority classes in the dataset without ignoring them. 

% \begin{figure}
%     \centering
%     \includegraphics[width=0.4\textwidth]{analysis_plan/memotion_matrix.png}
%     \caption{Confusion matrix for the best model for Memotion 7k dataset}
%     \label{memotion_matrix}
% \end{figure}
%\FloatBarrier

% From Figure \ref{memotion_matrix}, we find that apart from the samples correctly predicted on the diagonal, the biggest issue is that 154 neutral labels are misclassified as positive. We further delved into the index of these 154 misclassified labels in the test set and found some neutral, harmless expressions or statements being wrongly identified as positive because they don't exhibit any negative emotions. 
\begin{figure}[htbp]
  \centering
  \begin{minipage}[c]{0.45\textwidth}
    \centering
    \includegraphics[width=0.6\linewidth]{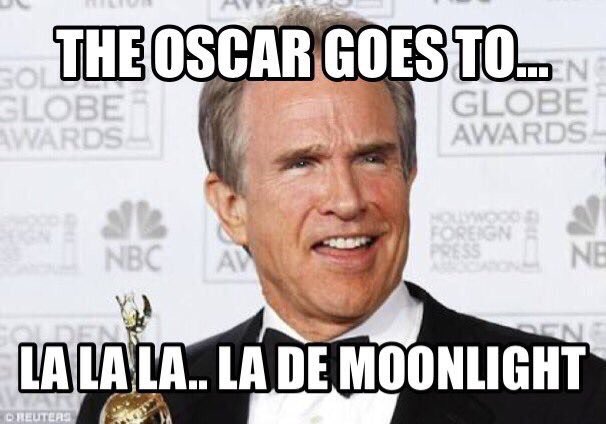}
    \caption*{Example 1}
    \label{fig:sample2}
  \end{minipage}
  \hfill
  \begin{minipage}[c]{0.45\textwidth}
    \centering
    \includegraphics[width=0.6\linewidth]{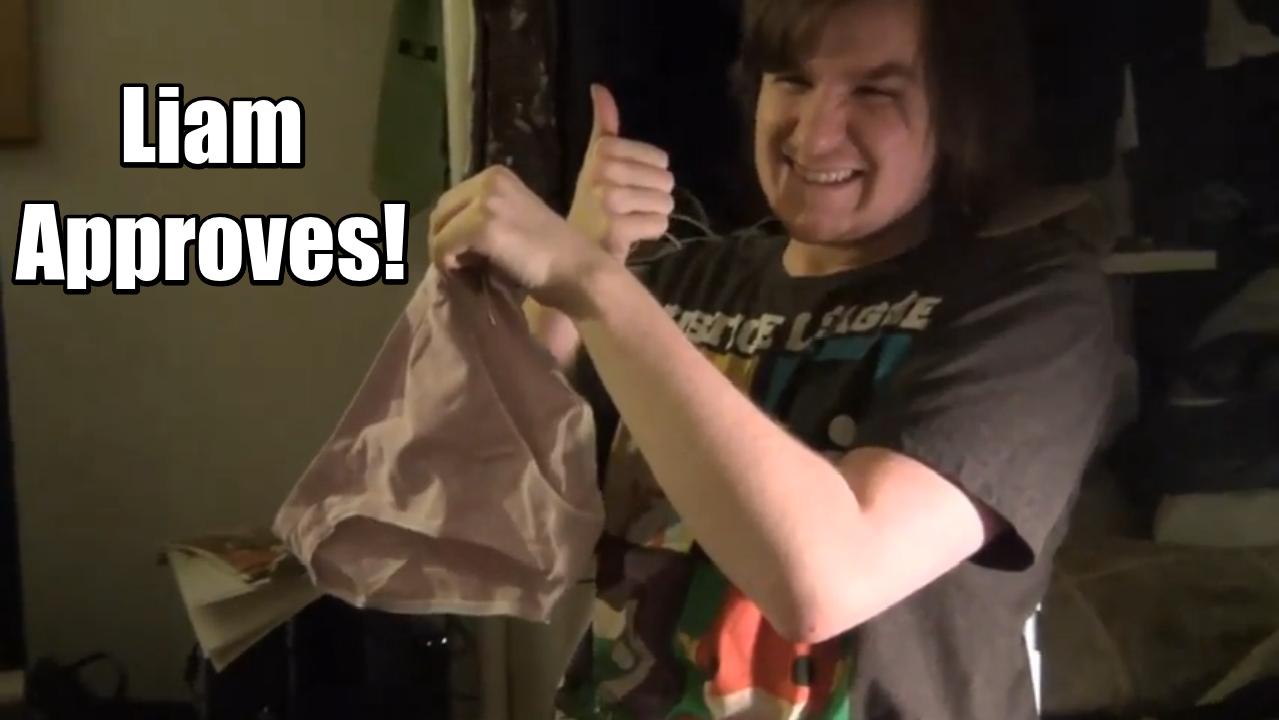}
    \caption*{Example 2}
    \label{fig:sample3}
  \end{minipage}
  
  \caption{Example of misclassified labels}
  \label{fig:misclassified_labels}
\end{figure}

Further, we selected two fairly typical photos from the set and found that all of them feature individuals who are smiling. This will likely make the model perceive them as expressing positive emotions (as shown in Figure \ref{fig:misclassified_labels}). However, another possible reason for the model's incorrect prediction could be issues with the dataset's annotation. As can be seen from Example 1 in Figure \ref{fig:misclassified_labels}, although the model has classified a 'neutral' label as 'positive', when we consider what he is saying, from a human perspective, this should be a 'negative' class because his smile is dry. By using image embeddings that include text, a more nuanced representation of the visual and textual data can be captured, which will help improve the accuracy of emotion classification. This approach will enhance the model's ability to discern more subtle emotional expressions and contextual factors that influence perceived emotions, leading to more accurate predictions.

\section{Conclusion}
Our model shows outstanding performance over multiple datasets, outperforming both baseline and exceeding (or reaching) state-of-the-art performance. This indicates that the model we proposed, along with the three fusion methods, can be well suited to multimodal sentiment analysis. In all three datasets, each fusion strategy yielded a best-performing model, indicating that our fusion methods are highly adept at integrating mutual information across multiple modalities. Although our model's performance is close to or surpasses current SOTA models, there are areas for future improvement. In the future, we will incorporate additional modal inputs like audio and explore advanced fusion techniques and different feature extraction models, including variants of BERT and CNN frameworks. 
 
\section*{Acknowledgment}
We acknowledge the contributions of Yufan Lin to this project.

\bibliographystyle{splncs04}
\bibliography{sample-base}

\end{document}